\pdfoutput=1
\documentclass[11pt]{article}

\usepackage[]{emnlp2021}

\usepackage{times}
\usepackage{latexsym}

\usepackage[T1]{fontenc}
\usepackage[utf8]{inputenc}

\usepackage{microtype}
\usepackage{amsmath}
\newcommand{\argmax}{\operatornamewithlimits{argmax}}

\usepackage{multirow}
\usepackage{multicol}
\usepackage{makecell}
\usepackage{graphicx} 
\usepackage{amssymb}
\usepackage{algorithm}
\usepackage{algpseudocode}
\usepackage{ulem}
\usepackage{booktabs}
\usepackage{CJKutf8}
\usepackage{subfigure}
\newcommand{\tabincell}[2]{\begin{tabular}{@{}#1@{}}#2\end{tabular}}
\usepackage{arydshln}

\title{Mixup Decoding for Diverse Machine Translation}
\author{Jicheng Li$^1$$^2$\footnotemark[2], 
Pengzhi Gao$^3$, Xuanfu Wu$^1$$^2$, 
Yang Feng$^1$$^2$\footnotemark[1], 
Zhongjun He$^3$, 
\\ {\bf Hua Wu$^3$,} and {\bf Haifeng Wang}$^3$ \\
       $^{1}$ Key Laboratory of Intelligent Information Processing, \\ Institute of Computing Technology, Chinese Academy of Sciences (ICT/CAS) \\ 
       $^{2}$ University of Chinese Academy of Sciences\\
       $^{3}$ Baidu Inc. No. 10, Shangdi 10th Street, Beijing, 100085, China\\
       \texttt{\{lijicheng,wuxuanfu20s,fengyang\}@ict.ac.cn} \\
       \texttt{\{gaopengzhi,hezhongjun,wu\_hua,wanghaifeng\}@baidu.com} 
       }
\begin{document}
\maketitle

\renewcommand{\thefootnote}{\fnsymbol{footnote}}
\footnotetext[2]{This work was done when Jicheng Li was interning at Baidu Inc., China.}
\footnotetext[1]{Yang Feng is the corresponding author of the paper.}

\begin{abstract}
Diverse machine translation aims at generating various target language translations for a given source language sentence. To leverage the linear relationship in the sentence latent space introduced by the mixup training, we propose a novel method, \textit{MixDiversity}, to generate different translations for the input sentence by linearly interpolating it with different sentence pairs sampled from the training corpus during decoding. To further improve the faithfulness and diversity of the translations, we propose two simple but effective approaches to select diverse sentence pairs in the training corpus and adjust the interpolation weight for each pair correspondingly. Moreover, by controlling the interpolation weight, our method can achieve the trade-off between faithfulness and diversity without any additional training, which is required in most of the previous methods. Experiments on WMT'16 \texttt{en}$\rightarrow$\texttt{ro}, WMT'14 \texttt{en}$\rightarrow$\texttt{de}, and WMT'17 \texttt{zh}$\rightarrow$\texttt{en} are conducted to show that our method substantially outperforms all previous diverse machine translation methods.
\end{abstract}

\section{Introduction} \label{sec:int}

Neural machine translation (NMT) \citep{NIPS2014_a14ac55a,wu2016google,gehring2017convolutional,NIPS2017_3f5ee243,ott2018scaling} has achieved significant success in improving the quality of machine translation. Despite these successes, NMT still faces problems in translation diversity \citep{vanmassenhove2019lost,gu2020token}. Due to the existence of lexical diversity, syntactic diversity and synonymous words in the target language, one source language sentence usually corresponds to multiple proper translations. However, existing NMT models mostly consider the one-to-one mapping but neglects the one-to-many mapping between the source and target languages.

Many studies have been proposed to tackle such issues by exploiting the diversity in the model space, such as using different experts \cite{shen2019mixture}, applying different multi-head attentions \cite{sun2020multihead}, and utilizing different models \cite{wu2020condropout}. Although the model-oriented methods have been well studied, the data-oriented method still lacks exploration.

\begin{figure}[!t]
\centering
\includegraphics[width=1\columnwidth]{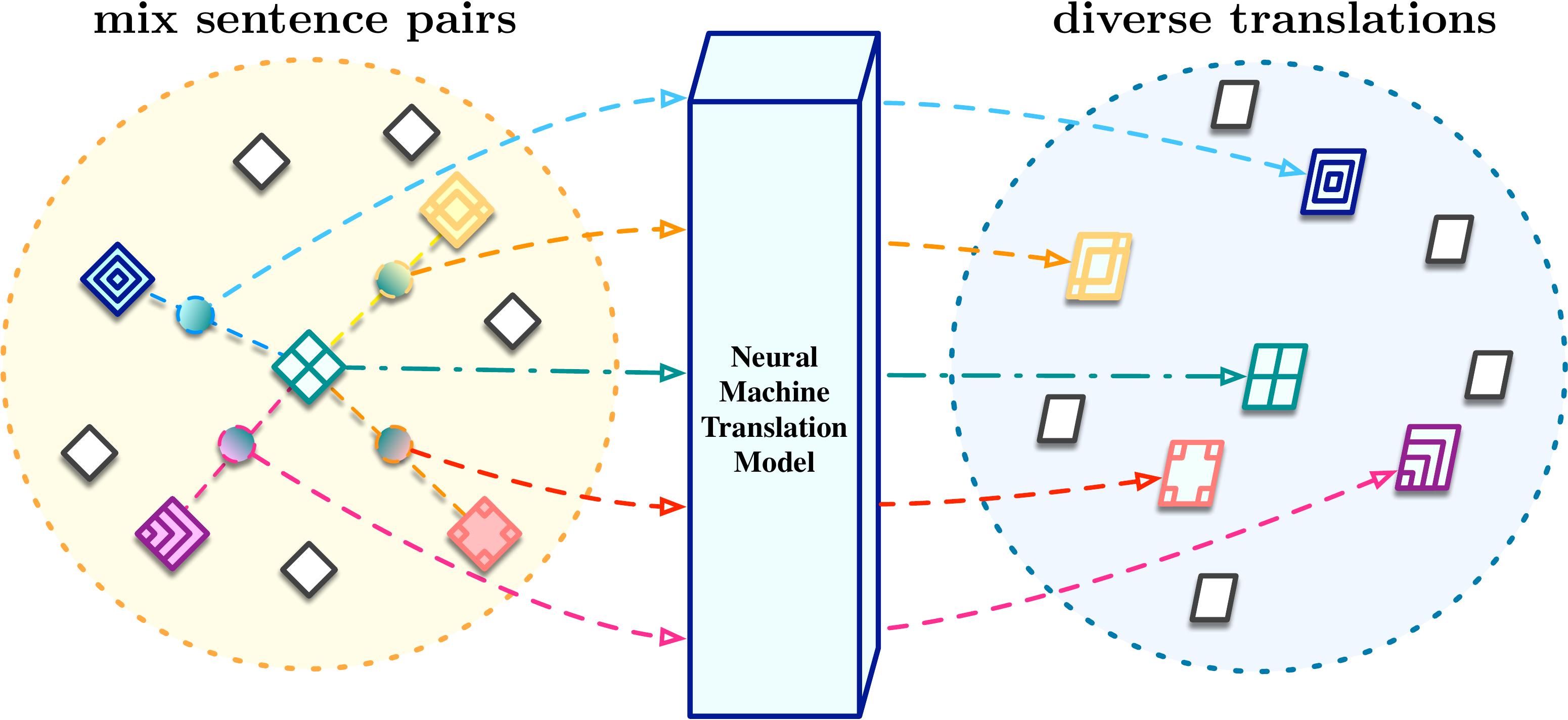}
\caption{Illustration of the proposed method, \textit{MixDiversity}, which linearly interpolates the input sentence with various sentence pairs sampled from the training corpus so as to generate diverse translations.} 
\label{fig:1} 
\vspace{-1em}
\end{figure} 

In this work, we focus on improving the translation diversity by exploiting the diversity in the sentence space. Since different translations of one source sentence share the same semantics, their sentence-level embeddings will gather in the same region in the target sentence space. In other words, each sentence in this region is a translation of the source sentence. By sampling different sentences from this region, we can obtain various translations. To sample different translations from this region, we propose a simple but effective method, \textit{MixDiversity}. As aforementioned, the NMT model learns a one-to-one mapping between the source and target languages. Given the source sentence and the generated tokens in the decoder, the NMT model can map the source sentence into a corresponding target sentence. Therefore, to obtain various translations on the target side, we need to find the corresponding inputs for the NMT model. By mixing the source sentence with the sampled sentence pairs in the training corpus via linear interpolation, we can obtain mixed sentences as inputs for the NMT model and map them into a corresponding sentence in the target sentence space. By assigning a larger interpolation weight for the source sentence, the mixed sentence then has similar semantics, and the corresponding translation has higher faithfulness to the source sentence. In this way, by mixing the source sentence with different sentence pairs during decoding, we can obtain diverse mixed sentences as inputs for the NMT model and map them to different translations for the source sentence.
 
Given that NMT models are non-linear functions, the interpolation weight of the input sentences could decline, and the semantic of the output could shift to the randomly sampled sentence pairs. To guarantee the consistency of the interpolation weight during decoding, we force the NMT model to learn to maintain the proportion between the mixed sentences with the mixup training strategy \citep{guo2020seqmix} which linearly interpolates two randomly sampled sentence pairs in both encoder and decoder during training. The main idea of our approach is illustrated in Figure \ref{fig:1}, where we mix one source sentence with four different sentence pairs sampled from the training corpus to obtain four variant mixed samples as inputs for the NMT model and map the mixed sentences to four diverse sentences in the target space.

\section{MixDiversity}

\subsection{Overview}
During training, we linearly interpolate word embeddings of two randomly sampled sentence pairs on both the source and target sides. During inference, since the corresponding target sentence of the input can not be obtained in advance, we interpolate word embeddings of previously generated tokens and the sampled target sentence in the decoder. Note that the MixDiversity can also be used without the Mixup Training.

\subsection{Mixup Training for NMT}

We apply the mixup training \citep{guo2020seqmix} to encourage the NMT model to learn the linear relationship in the latent space of the input sentences. %Given a parallel corpus $\mathcal{S} = \{\mathbf{x}^i, \mathbf{y}^i\}_{i=1}^{|\mathcal{S}|}$, 
Consider a pair of training samples $(\mathbf{x}^i, \mathbf{y}^i)$ and $(\mathbf{x}^j, \mathbf{y}^j)$ in the parallel corpus $\mathcal{S}$, where $\mathbf{x}^i$ and $\mathbf{x}^j$ denote the source sentences, and $\mathbf{y}^i$ and $\mathbf{y}^j$ denote the target sentences. The synthetic sample $(\mathbf{x}^{ij}, \mathbf{y}^{ij})$ is generated as follows.
\begin{equation}\nonumber
\mathbf{x}^{ij} = \lambda \mathbf{x}^i + (1 - \lambda) \mathbf{x}^j, \ \ \mathbf{y}^{ij} = \lambda \mathbf{y}^i + (1 - \lambda) \mathbf{y}^j,
\end{equation}
where $\lambda$ is drawn from a Beta distribution $\text{Beta}(\alpha, \alpha)$ with a hyper-parameter $\alpha$. 
The synthetic sample $(\mathbf{x}^{ij}, \mathbf{y}^{ij})$ is then fed into the NMT model for training to minimize the empirical risk:
\begin{equation}\label{mixup-loss}
\mathcal{L}(\theta) = \mathop{\mathbb{E}}\limits_{\substack{(\mathbf{x}^i, \mathbf{y}^i) \in \mathcal{S} \\ (\mathbf{x}^j, \mathbf{y}^j) \in \mathcal{S}}} [\ell(f(\mathbf{x}^{ij}, \mathbf{y}^{ij}; \theta), \ddot{\mathbf{y}}^{ij})],
\end{equation}
where $\ell$ denotes the cross entropy loss,  $\theta$ is a set of model parameters, $f(*)$ is the probability predictions of the NMT model, 
\begin{equation}
\ddot{\mathbf{y}}^{ij} = \lambda \ddot{\mathbf{y}}^{i} + (1 - \lambda) \ddot{\mathbf{y}}^{j},
\end{equation}
and $\ddot{\mathbf{y}}^{i}$ and $\ddot{\mathbf{y}}^{j}$ are the sequences of one-hot label vectors for $\mathbf{y}^i$ and $\mathbf{y}^j$ respectively.

\subsection{Mixup Decoding for Diverse MT}

At inference, assume $\mathbf{x} = x_1, ..., x_I$ that corresponds to the source sentence with length $I$. We mix it with $K$ different sentence pairs $(\mathbf{x}^1, \mathbf{y}^1),\ldots,(\mathbf{x}^K, \mathbf{y}^K)$ selected from the training corpus to generate $K$ different translations of $\mathbf{x}$. Specifically, for the $i^{th}$ translation, we first interpolate the token embeddings of $\mathbf{x}$ with the token embeddings of $\mathbf{x}^i$ in the encoder side:
\begin{equation}\label{encoder}
\hat{e}(x^i_t) = \lambda_{t}^{i} e(x_t) + (1 - \lambda_{t}^{i}) e(x^i_t), \ \ \forall t \in [1, I].
\end{equation}
The encoder then maps the mixed token embeddings $\hat{e}(\mathbf{x}^i_1), \ldots \hat{e}({x}^i_I)$ into the corresponding hidden representations $\mathbf{h}^i$.

In the decoder side, at step $t$, we mix the embedding of the token ${y}_{t-1}$, which is predicted by the NMT model at step $t-1$, with the embedding of ${y}^i_{t-1}$ as follows:
\begin{equation}\label{decoder}
\hat{e}({y}^i_{t-1}) = \lambda_{t}^{i} e({y}_{t-1}) + (1 - \lambda_{t}^{i}) e({y}^i_{t-1}),
\end{equation}
where ${y}_{0}$ and ${y}^{i}_{0}$ are the special beginning-of-sentence symbol $\langle bos \rangle$. The predicted token ${y}_t$ is then calculated by
\begin{equation}
{y}_t = \argmax_{y \in {\mathcal{V}_{{y}}}} P(y|\mathbf{h}^i, \hat{e}({y}^i_{\leqslant t-1}); \theta), \; t\geqslant 1,
\end{equation}
where ${\mathcal{V}_{{y}}}$ is the vocabulary of the target language. Note that $\lambda_{t}^{i}$'s in \eqref{encoder} and \eqref{decoder} are drawn from the Beta distribution $\text{Beta}(\alpha, \alpha)$ with the same $\alpha$ for different $t$ and $i$. 

\paragraph{Select Sentence Pairs by Source Length} 

We first group sentence pairs in the training corpus by their source sentence lengths and then randomly select $K$ sentence pairs $(\mathbf{x}^1, \mathbf{y}^1), \ldots (\mathbf{x}^K, \mathbf{y}^K)$ from the groups that have similar length compared with the input sentence. Specifically, given an input sentence with length $I$, we sample sentence pairs from the groups with lengths in the range of $[I-1, I]$.

\paragraph{Adjust Interpolation Weight by Similarity}

In order to correctly translate the semantic of the input sentence, $\mathbf{x}$ needs to dominate the mixed samples. Different sentences in $(\mathbf{x}^1, \mathbf{y}^1), \ldots (\mathbf{x}^K, \mathbf{y}^K)$ may have different similarity with $\mathbf{x}$, and a higher similarity between $\mathbf{x}^{i}$ and $\mathbf{x}$ implies a looser constraint on the interpolation weight between them. Thus, taking the similarity between $\mathbf{x}^{i}$ and $\mathbf{x}$ into account, we sample the interpolation weight $\lambda_{t}^{i}$ from the Beta distribution as follows. 
\begin{equation} \label{eq:3}
\lambda_{t}^{i} \sim \text{Beta}(\alpha_{i},\alpha_{i}), \ \ \alpha_{i} = \tau + \frac{\tau}{d(\mathbf{x},\mathbf{x}^{i})}, 
\end{equation}
where $\tau$ is a hyper-parameter to control the interpolation weight, and $d(*)$ is the Euclidean distance between the embeddings of two sentences, which are defined as the average among all token embeddings in the sentence. In our implementation, $\lambda_{t}^t$ is actually set to be $\max(\lambda_{t}^t, 1 - \lambda_{t}^t)$. The larger distance between $\mathbf{x}$ and $\mathbf{x}^i$ is, the larger interpolation weight $\lambda_{t}^t$ we have, which leads to dynamically adjusting on the interpolation weight based on the sentence similarity.

\section{Experimental Setup}

\subsection{Data Description}

Our experiments consider three translation datasets: WMT'16 English-Romanian (\texttt{en}$\rightarrow$\texttt{ro}), WMT'14 English-German (\texttt{en}$\rightarrow$\texttt{de}), and WMT'17 Chinese-English (\texttt{zh}$\rightarrow$\texttt{en}). All sentences are prepossessed with byte-pair-encoding (BPE) \citep{sennrich-etal-2016-edinburgh}. For WMT'16 \texttt{en}$\rightarrow$\texttt{ro}, we use the preprocessed dataset released in \citet{lee-etal-2018-deterministic} which contains 0.6M sentence pairs. We use newsdev-2016 as the validation set and newstest-2016 as the test set. We build a shared vocabulary with 40K BPE types. For WMT'14 \texttt{en}$\rightarrow$\texttt{de}, it consists of 4.5M training sentence pairs, and we use newstest-2013 for validation and newstest-2014 for test. We build a shared vocabulary with 32K BPE types. For WMT'17 \texttt{zh}$\rightarrow$\texttt{en}, it consists of 20.1M training sentence pairs, and we use devtest-2017 as the validation set and newstest-2017 as the test set. We build the source and target vocabularies with 32K BPE types separately.

\subsection{Model Configuration}

We apply a standard 6-layer Transformer Base model \citep{NIPS2017_3f5ee243} with 8 attention heads, embedding size 512, and FFN layer dimension 2048. We use the label smoothing \citep{Szegedy2016labelsmooth} with $\epsilon=0.1$ and Adam \citep{kingma2017adam} optimizer with $\beta_{1}=0.9$, $\beta_{2}=0.98$ and $\epsilon=10^{-9}$. We set learning rate as $0.0007$ with 4000 warmup steps from the initialized learning rate of $10^{-7}$. The NMT model is trained with dropout 0.1 and max tokens 4096. When adopting the mixup training strategy, we set $\alpha$ as 1.0, 0.1 and 0.1 for \texttt{en}$\rightarrow$\texttt{ro}, \texttt{en}$\rightarrow$\texttt{de} and \texttt{zh}$\rightarrow$\texttt{en} respectively. We train our model on 4 NVIDIA V100 GPUs until it converges. At the inference time, we set beam size as 4 with length penalty 0.6.

\begin{table}[!t]
\centering
\small
\begin{tabular}{lccc}
\toprule
\multicolumn{1}{c}{\multirow{2}{*}{Strategy}} & \multicolumn{3}{c}{Baseline BLEU $\mathcal{R}$} \\
\cmidrule(lr){2-4}
&\texttt{en}$\rightarrow$\texttt{ro} & \texttt{en}$\rightarrow$\texttt{de} & \texttt{zh}$\rightarrow$\texttt{en} \\
\midrule
Vanilla  & 32.80 & 27.43 & 24.07 \\
Mixup  & 33.75 & 27.70 & 24.40 \\
\bottomrule
\end{tabular}
\caption{The baseline BLEU of different training strategy in each dataset.}
\label{tab:eda}
\vspace{-1em}
\end{table}

\begin{table*}[!th]
\centering
\small
\begin{tabular}{lccccccccc}
\toprule
\multicolumn{1}{c}{\multirow{3}{*}{Method}} & \multicolumn{3}{c}{WMT'16 \texttt{en}$\rightarrow$\texttt{ro}} & \multicolumn{3}{c}{WMT'14 \texttt{en}$\rightarrow$\texttt{de}} & \multicolumn{3}{c}{WMT'17 \texttt{zh}$\rightarrow$\texttt{en}} \\
\cmidrule(lr){2-4} \cmidrule(lr){5-7} \cmidrule(lr){8-10}
& \text{rfb}$\boldsymbol\Uparrow$ & \text{pwb}$\boldsymbol\Downarrow$ & \text{EDA}$\boldsymbol\Downarrow$ & \text{rfb}$\boldsymbol\Uparrow$ & \text{pwb}$\boldsymbol\Downarrow$ & \text{EDA}$\boldsymbol\Downarrow$ & \text{rfb}$\boldsymbol\Uparrow$ & \text{pwb}$\boldsymbol\Downarrow$ & \text{EDA}$\boldsymbol\Downarrow$ \\
\midrule
BeamSearch (BS) & 31.99 & 80.82 & 26.62 & 26.51 & 77.61 & 21.55 & 23.69 & 81.36 & 19.64 \\
\midrule
DiverseBS \citep{vijayakumar2018diverse} & 30.65 & 76.46 & 25.92 & 24.78 & 66.81 & 20.71 & 22.43 & 66.93 & 17.49 \\
HardMoE \citep{shen2019mixture} & 31.13 & 68.42 & 23.01 & 23.68 & \textbf{51.69} & 19.69 & 21.77 & \textbf{49.13} & 15.20 \\
HeadSample \citep{sun2020multihead} & 26.94 & \textbf{59.38} & 26.42 & 25.05 & 76.55 & 22.71 & 21.24 & 73.96 & 21.33 \\
ConcreteDropout \citep{wu2020condropout} & 31.20 & 65.24 & 21.94 & 25.12 & 60.02 & 18.49 & \textbf{23.10} & 55.61 & 13.97 \\
\hdashline
MixDiversity & \textbf{31.50} & 59.57 & \textbf{21.10} & \textbf{25.50} & 57.50 & \textbf{17.79} & 22.96 & 51.52 & \textbf{13.88} \\
\quad \textit{w/o} Mixup Training & 31.48 & 66.44 & 22.16 & 25.24 & 59.43 & 18.15 & 22.87 & 54.01 & 13.92 \\
\bottomrule
\end{tabular}
\caption{The best result of each method on WMT'16 \texttt{en}$\rightarrow$\texttt{ro}, WMT'14 \texttt{en}$\rightarrow$\texttt{de}, and WMT'17 \texttt{zh}$\rightarrow$\texttt{en}. For DiverseBS, HardMoE, and HeadSample, we select the result under the best settings described in their papers. For ConcreteDropout and MixDiversity, we validate the model under different hyper-parameter settings on the validation set to find the best settings for the model, and we report the result on the test set under the best settings. We get the best results of MixDiversity with $\tau=0.3$, $0.3$, and $0.25$ in \texttt{en}$\rightarrow$\texttt{ro}, \texttt{en}$\rightarrow$\texttt{de} and \texttt{zh}$\rightarrow$\texttt{en} respectively. $\boldsymbol\Uparrow$ means the higher, the better. $\boldsymbol\Downarrow$ means the lower, the better.}
\label{tab:1}
\vspace{-0.5em}
\end{table*}

\begin{figure}[!t]
\centering
\includegraphics[width=0.85\columnwidth]{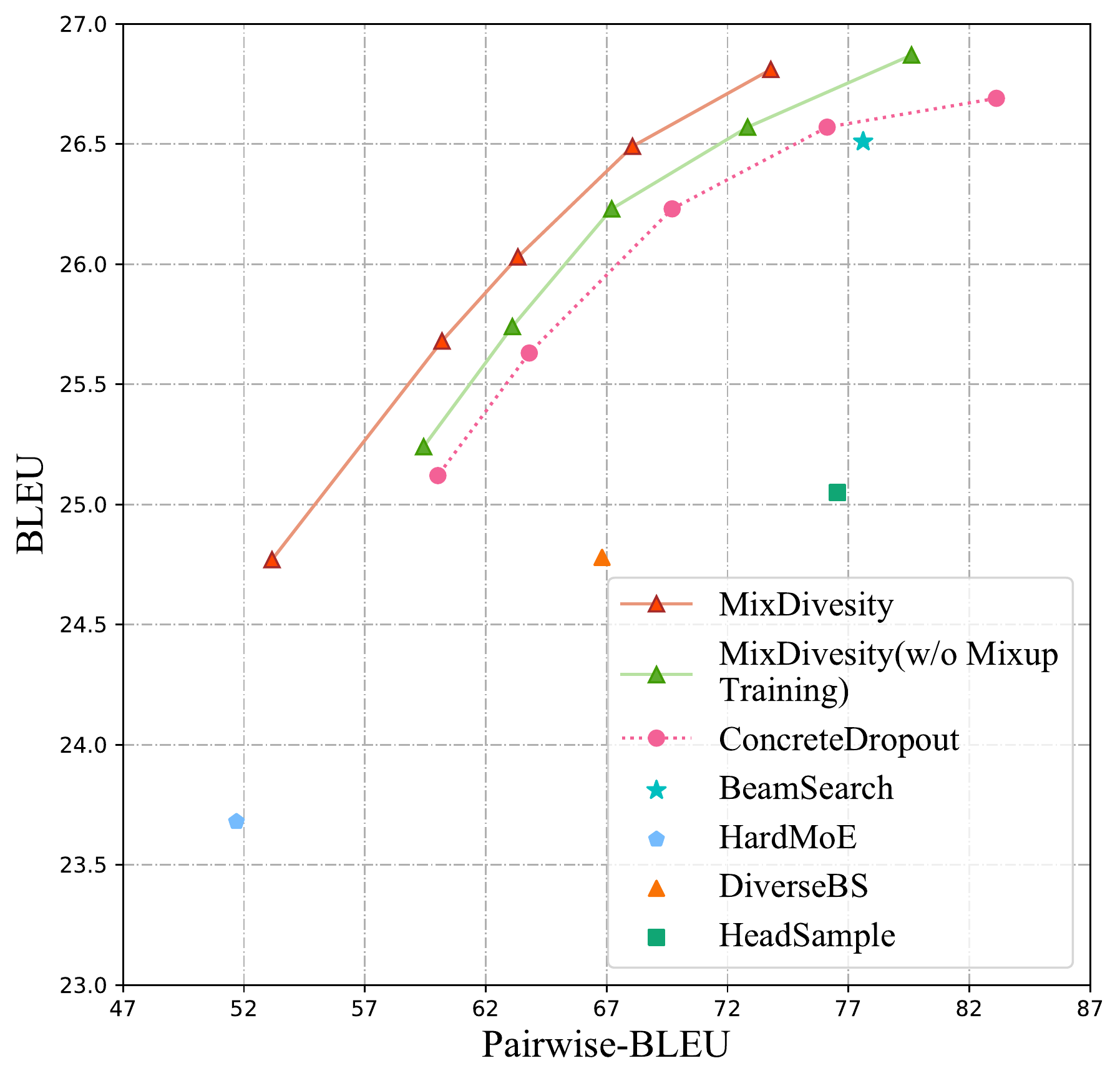}
\caption{Illustration of the trade-off between reference BLEU and pair-wise BLEU in WMT'14 \texttt{en}$\rightarrow$\texttt{de} with different $\tau$.} 
\label{fig:ende} 
\vspace{-0.5em}
\end{figure}

\subsection{Evaluation Metrics}

Referring to \citet{wu2020condropout}, we adopt the average BLEU with reference (rfb) to measure the faithfulness of different translations to the input sentence and the average pairwise-BLEU (pwb) to measure the pair-wise similarity between different translations. The higher rfb, the better accuracy of the translations. The lower pwb, the better diversity of the translations. In our experiments, given one input sentence, we generate five different translations for all methods.

When we calculate Diversity Enhancement per Quality (DEQ) \citep{sun2020multihead} to evaluate the overall performance of different methods, we find that the DEQ results are not stable. For instance, the DEQ scores of ConcreteDropout in Figure~\ref{fig:ende} (from the leftmost point to the rightmost point) are 12.65, 15.69, 28.21, -24.83, and 30.61, where positive and negative scores appear alternately. We thus propose a new metric, Euclidean Distance from the ultimate Aim (EDA), to evaluate the overall quality of the results synthetically.

Consider rfb and pwb as the abscissa and the ordinate of a coordinate system, where $0 \leqslant \text{rfb} \leqslant \mathcal{R}$, and $0 \leqslant \text{pwb} \leqslant \mathcal{P}$. $\mathcal{R}$ is the baseline BLEU, which is defined as the BLEU score of the top one translation by beam search decoding with beam size 4 in our experiments. $\mathcal{P}=100$ is the maximal pwb. Different results with specific rfb and pwb scores could be mapped to different points in this coordinate system. The ultimate aim of the diverse machine translation task is to reach the point $(\mathcal{R},0)$. By measuring the Euclidean distance between $(\mathcal{R},0)$ and the result, we can evaluate the overall quality of the result. 

We, however, notice that rfb and pwb have different ranges ($\mathcal{P} > \mathcal{R}$), and pwb decreases much faster than rfb with the changing of $\tau$. As a consequence, the calculated EDA is biased to the results with the lower pwb scores. To alleviate such bias, we normalize the value of rfb and pwb to $[0, 1]$ by dividing $\mathcal{R}$ and $\mathcal{P}$ respectively and add a weight $\omega=\frac{\mathcal{R}}{\mathcal{P}}$ on the pwb term shown as follows:
\begin{equation}\label{eq:eda}\nonumber
\text{EDA} = 100\% \cdot \sqrt{(\frac{\mathcal{R}-\text{rfb}}{\mathcal{R}}{)}^{2}+{\omega}^{2}(\frac{0-\text{pwb}}{\mathcal{P}}{)}^{2}}.
\end{equation}
Note that different training strategies lead to different baseline BLEU $\mathcal{R}$. Table \ref{tab:eda} shows the baseline BLEU of Transformer in each dataset. When we use EDA to evaluate the performance of ConcreteDropout in Figure~\ref{fig:ende}, we get 18.49, 18.69, 19.61, 21.1, and 22.95. This result shows that EDA is a better and more stable overall evaluation metric than DEQ for the diverse machine translation.

\section{Experimental Results}

\subsection{Main Results}

We show the results of different methods on generating diverse translations in Table \ref{tab:1}. We compare our method with the conventional beam search decoding (BeamSearch) and the existing model-oriented methods, including DiverseBS, HardMoE, HeadSample, and ConcreteDropout. For each method, we exhibit its best result with the lowest EDA score. We can see that MixDiversity gets lower EDA scores than all existing methods in all three datasets, and the performance of MixDiversity without the mixup training also outperforms other competitors on WMT'14 \texttt{en}$\rightarrow$\texttt{de} and WMT'16 \texttt{zh}$\rightarrow$\texttt{en} with lower EDA scores. 

Figure \ref{fig:ende} shows the trade-off results between the reference BLEU and the pair-wise BLEU on WMT'14 \texttt{en}$\rightarrow$\texttt{de}. We can see that mixup training or not, MixDiversity generally performs better than all other methods without additional training or finetuning, which is required in most previous methods, such as HardMoE.

\subsection{Ablation Study} \label{sec:ab}

The results of the ablation study are shown in Table \ref{tab:ablation}, which consists of three experiments. In the first experiment, we evaluate the performance of our method with different settings: training NMT models without mixup strategy (\textit{w/o} Mixup Training), decoding by randomly selecting $K$ sentence pairs from the entire training corpus (\textit{w/o} LenSelection), and sampling the interpolation weights without considering similarities between $\mathbf{x}$ and $\mathbf{x}^{i}$ (\textit{w/o} SimWeight). In the second experiment, we not only attempt to mix the input sentence with Gaussian noise drawn from $\mathcal{N}(0, 2)$, but we also mix the input sentence with synthetic sentence pairs which are made up of tokens that are randomly sampled from the vocabulary. In both cases, we observe remarkable increases in EDA. Such a phenomenon indicates that the potential linguistic features in training samples could assist MixDiversity in generating different translations of high diversity and faithfulness. In the last experiment, we verify the rationality and effectiveness of the mixup operations in both encoder and decoder. 

\begin{table}
\centering
\small
\begin{tabular}{llccc}
\toprule
 & \multicolumn{1}{c}{Method} & \text{rfb}$\boldsymbol\Uparrow$ & \text{pwb}$\boldsymbol\Downarrow$ & \text{EDA}$\boldsymbol\Downarrow$ \\
\midrule
\multicolumn{1}{c}{\multirow{4}{*}{1}} & MixDiversity ($\tau=0.3$) & 25.50 & 57.50 & \textbf{17.79}  \\
&\quad \textit{w/o} Mixup Training & 25.24 & 59.43 & 18.15  \\
&\quad \textit{w/o} LenSelection & 25.58 & 65.13 & 19.09  \\
&\quad \textit{w/o} SimWeight & 25.77 & 69.99 & 20.12  \\
\midrule
\multicolumn{1}{c}{\multirow{4}{*}{2}} & MixDiversity ($\tau=0.3$) &  &  &   \\
& \quad + Mixup Samples & 25.50 & 57.50 & \textbf{17.79}  \\
& \quad + Mixup Noises & 21.58 & 43.78 & 25.20 \\
& \quad + Mixup SynSents & 11.44 & 13.30 & 58.81  \\
\midrule
\multicolumn{1}{c}{\multirow{4}{*}{3}} & MixDiversity ($\tau=0.3$) &  &  &   \\
& \quad + Both Sides Mixup & 25.50 & 57.50 & \textbf{17.79}  \\
& \quad + Only Encoder Mixup & 25.83 & 66.09 & 19.51 \\
& \quad + Only Decoder Mixup & 25.41 & 60.69 & 18.73  \\
\bottomrule
\end{tabular}
\caption{Ablation study on WMT'14 \texttt{en}$\rightarrow$\texttt{de}.}
\label{tab:ablation}
% \vspace{-0.em}
\end{table}

\subsection{Applications of Diverse Translation}

In Table \ref{tab:app}, we compare MixDiversity with BeamSearch (BS) to show the application of diverse translation methods on boosting the performance of both Back Translation and Knowledge Distillation. We generate sentences with a beam size of 5 for all methods. For BeamSearch (Top 5) and MixDiversity, we generate five different translations. In the Back Translation experiment, we randomly sample 4M sentences from the German monolingual corpus distributed in WMT'18 and combine the original parallel corpus with the back-translated parallel corpus to train the NMT model. In the Data Distillation experiment, we train the student NMT model with the generated sentences of the teacher NMT model.

\section{Related Work}

Many studies have been proposed to improve the translation diversity by exploiting the diversity in the model space. \citet{li2016simple} and \citet{vijayakumar2018diverse} adopt various regularization terms in the beam search decoding to encourage generating diverse outputs. \citet{HeHN18} generates different translations by incorporating condition signals of different models. \citet{shen2019mixture} proposes to training NMT models with the mixture of experts method and generates diverse translations using different latent variables of different experts. \citet{shu2019diversecodes} generates diverse translation conditioned on different sentence codes. \citet{sun2020multihead} discovers that encoder-decoder multi-head attention in Transformer learns multiple target-source alignments and generates diverse translations by sampling different heads in the attention modules. \citet{wu2020condropout} samples different models from a posterior model distribution and employs variational inference to control the diversity of translations.

\begin{table}[!t]
\centering
\small
\begin{tabular}{lcc}
\toprule
& \text{Back Trans.} & \text{Knowledge Distill.} \\
\midrule
Baseline & 27.43 & -- \\
\hdashline
BS (Top 1) & 28.81 & 27.28 \\
BS (Top 5) & 28.82 & 27.46 \\
MixDiversity & \textbf{29.19} & \textbf{27.83} \\
\bottomrule
\end{tabular}
\caption{Results of the Back Translation and the Knowledge Distillation experiments on WMT'14 \texttt{en}$\rightarrow$\texttt{de}.}
\label{tab:app}
\vspace{-1.5em}
\end{table}

\section{Conclusion}

In this work, we propose a novel method, \textit{MixDiversity}, for the diverse machine translation. Compared with the previous model-oriented methods, MixDiversity is a data-oriented method that generates different translations of the input sentence by utilizing the diversity in the sentence latent space. We also propose two simple but effective methods to select the mixup samples and adjust the mixup weights for each sample. To evaluate the overall performance synthetically, we design a new evaluation metric, \textit{EDA}. Experimental results show that MixDiversity outperforms all previous methods in the field of diverse machine translation.

\section*{Acknowledgements}

We thank all the anonymous reviewers for their insightful and valuable comments. This work was supported by National Key R\&D
Program of China (NO. 2017YFE0192900).

\bibliography{anthology,custom}
\bibliographystyle{acl_natbib}

% \clearpage
\appendix

% \begin{figure}
% \centering
% \subfigure[\texttt{en}$\rightarrow$\texttt{ro}]{
% \includegraphics[scale=0.35]{enro.pdf}}
% \subfigure[\texttt{zh}$\rightarrow$\texttt{en}]{
% \includegraphics[scale=0.35]{zhen.pdf}}
% \caption{Illustration of the trade-off between reference BLEU and pair-wise BLEU in WMT'16 \texttt{en}$\rightarrow$\texttt{ro} and WMT'17 \texttt{zh}$\rightarrow$\texttt{en} with different $\tau$.}
% \label{fig:trade-off}
% \end{figure}

\section{Methods for Comparison}

In our experiments, we set $k=5$ and compare our method with the following works:

\begin{itemize}
\item \textbf{BeamSearch (BS)}: In our experiments, we choose the top $k$ sentences generated by beam search decoding as the result. 
\item \textbf{DiverseBeamSearch (DiveseBS)} \citep{vijayakumar2018diverse}: It generates diverse translations by grouping sentences in the beam search decoding with a regularization term to guarantee the diversity between different groups. We set the number of groups as $k$, and each group includes two sentences in our experiments.
\item \textbf{HardMoE} \citep{shen2019mixture}: It first trains the model with $k$ different hidden states and then generates different translations with different hidden states.
\item \textbf{HeadSample} \citep{sun2020multihead}: It generates different outputs by sampling different heads in multi-head attention modules. In our experiments, we set the number of heads to be sampled as 3.
\item \textbf{ConcreteDropout} \citep{wu2020condropout}: It generates different outputs by sampling different models from the model distribution using variational inference.
\end{itemize}

\section{Trade-off between reference BLEU and pair-wise BLEU }

\begin{figure}
\centering
\subfigure[\texttt{en}$\rightarrow$\texttt{ro}]{
\includegraphics[scale=0.35]{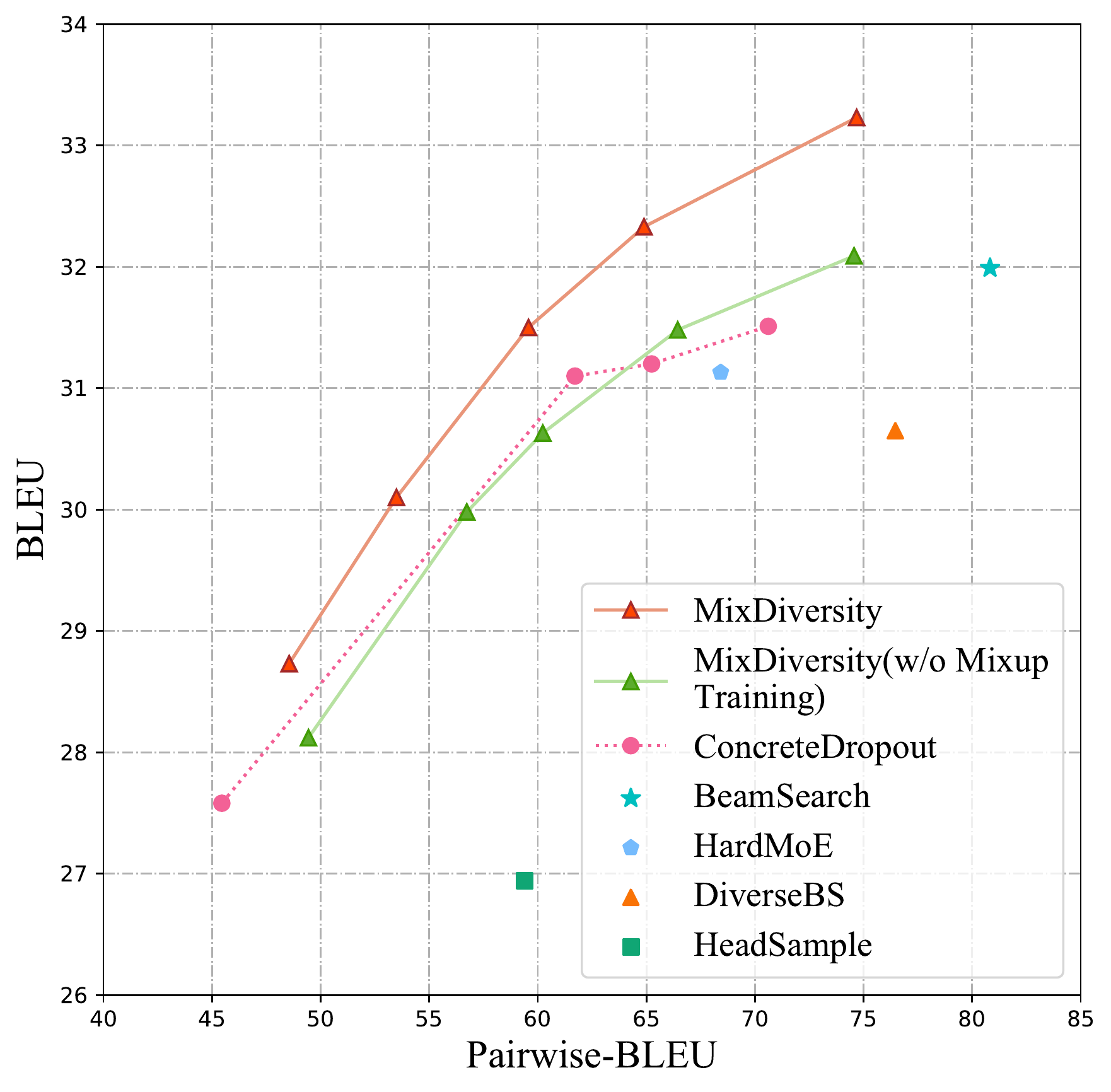}}
\subfigure[\texttt{zh}$\rightarrow$\texttt{en}]{
\includegraphics[scale=0.35]{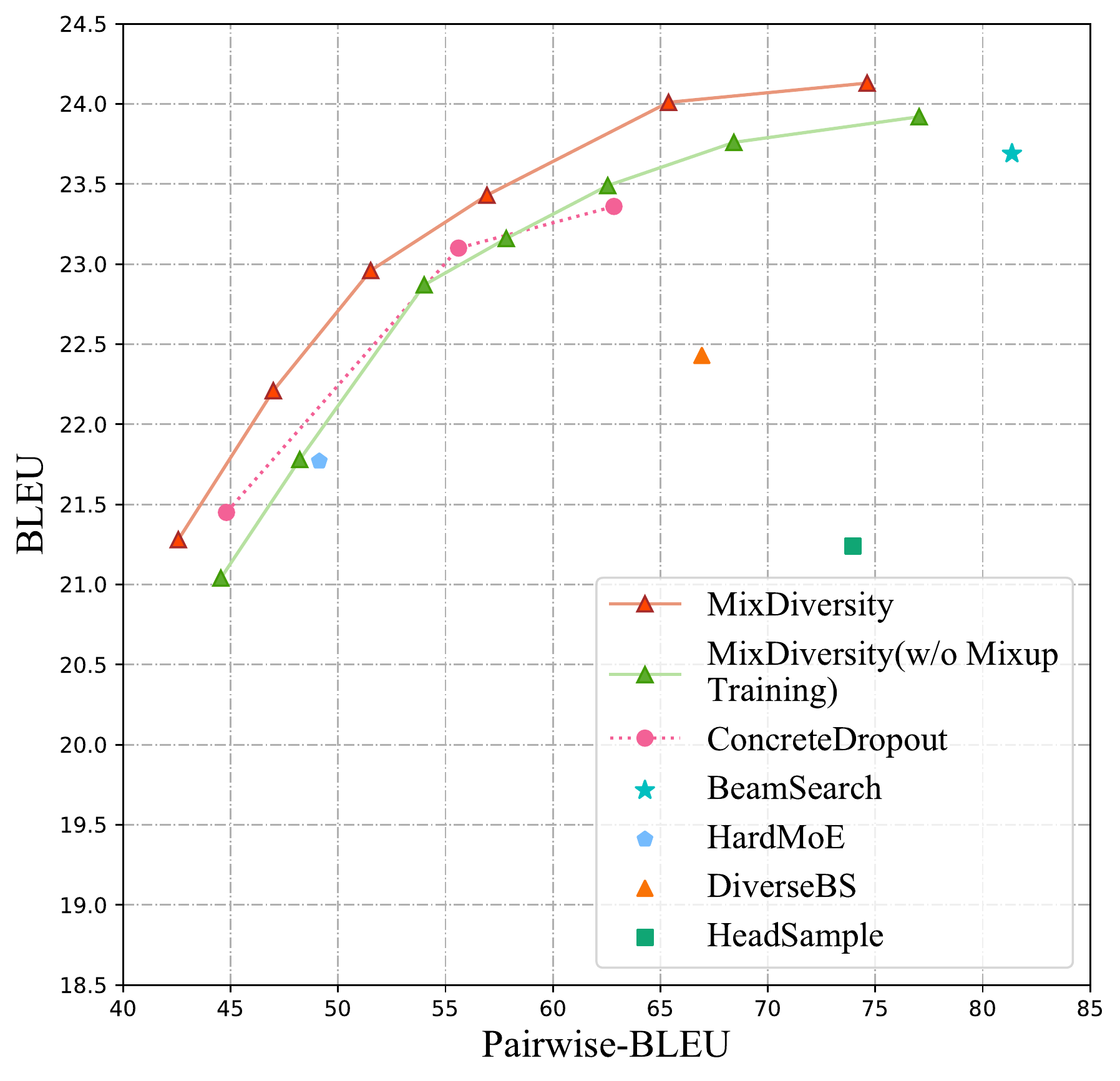}}
\caption{Illustration of the trade-off between reference BLEU and pair-wise BLEU in WMT'16 \texttt{en}$\rightarrow$\texttt{ro} and WMT'17 \texttt{zh}$\rightarrow$\texttt{en} with different $\tau$.}
\label{fig:trade-off}
\end{figure}

Figure \ref{fig:trade-off} shows the trade-off results between reference BLEU and pair-wise BLEU in WMT'16 \texttt{en}$\rightarrow$\texttt{ro} and WMT'17 \texttt{zh}$\rightarrow$\texttt{en}. From results in both \texttt{en}$\rightarrow$\texttt{ro} and \texttt{zh}$\rightarrow$\texttt{en}, we find that the lines of the MixDiversity and the ConcreteDropout overlap with each other. In addition, the ConcreteDropout needs to finetune the translation model under different configurations to achieve different trade-off results between the BLEU and the pair-wise BLEU. While the HardMoE needs to retrain the whole model with different settings of the number of experts so as to achieve the trade-off between the two BLEU scores. Besides, the performance of the HeadSample is unstable with different number of the sampled heads. In contrast, the MixDiversity can achieve the trade-off between the two BLEU scores by the hyper-parameter $\boldsymbol{\tau}$ without any additional training or finetuning time.

\begin{table*}[!ht]
    \centering
    \small
    \begin{tabular}{lccc}
    \toprule
     & {rf-BERTscore}$\boldsymbol\Uparrow$ & {pw-BERTscore}$\boldsymbol\Downarrow$ & EDA-BERTscore$\boldsymbol\Downarrow$\\
\midrule
Beam Search (BS) &\textbf{85.50} & 95.87  & 96.95\\
HeadSample \citep{sun2020multihead} & 84.99 & 96.29  & 97.45 \\
ConcreteDropout \citep{wu2020condropout} & 84.93 & 95.52  & 96.69 \\
\hdashline
MixDiversity (w/o Mixup Training) & 84.61 & \textbf{92.26} & \textbf{93.53} \\
    \bottomrule
    \end{tabular}
    \caption{The evaluation result using BERT-Score in WMT'14 \texttt{en}$\rightarrow$\texttt{de}. $\boldsymbol\Uparrow$ means the higher, the better. $\boldsymbol\Downarrow$ means the lower, the better.}
    \label{tab:bert-score}
\end{table*}

\begin{CJK}{UTF8}{gbsn}
\begin{table*}[!t]
    \centering
    \small
    \begin{tabular}{ll}
    \toprule
    Source & \tabincell{l}{因此 ， 人类 希望 有朝一日 在 火星 建立 居住 基地 ， 最终 向 火星 移民 ， 把 它 变成 人类 的\\ 第二 家园 。} \\
    \midrule
    Reference & \tabincell{l}{Therefore , the human beings hope that one day on the Mars to establish a base of residence ,\\ and ultimately to Mars immigration , it turned into a second home of mankind .} \\
    \midrule
     \multicolumn{1}{l}{\multirow{10}{*}{BeamSearch}} & \tabincell{l}{Therefore , human beings hope that one day they will establish a residence base on Mars and eventually \\emigrate to Mars , making it their second home .} \\
     \cdashline{2-2}[1pt/1pt]
    & \tabincell{l}{Therefore , the human race hopes one day to establish a residence base on Mars and eventually emigrate \\to Mars , making it the second home of the human race .} \\
    \cdashline{2-2}[1pt/1pt]
    & \tabincell{l}{Therefore , the human race hopes one day to establish a residence base on Mars and eventually emigrate \\to Mars , turning it into the second home of mankind .} \\
    \cdashline{2-2}[1pt/1pt]
    & \tabincell{l}{Therefore , human beings hope that one day they will establish a residence base on Mars and eventually \\emigrate to Mars , making it the second home of human beings .} \\
    \cdashline{2-2}[1pt/1pt]
    & \tabincell{l}{Therefore , human beings hope that one day they will establish a residence base on Mars and eventually \\emigrate to Mars , turning it into the second home of mankind .} \\
    \midrule
    \multicolumn{1}{l}{\multirow{10}{*}{\tabincell{c}{MixDiversity\\($\tau=0.15$)}}} & \tabincell{l}{Therefore , man hopes one day to establish a residence base on Mars , and eventually emigrate to \\Mars and turn it into a second home .}\\
    \cdashline{2-2}[1pt/1pt]
    & \tabincell{l}{So humans hope to one day establish a residence base on Mars and eventually emigrate to Mars and \\turn it into a second home for humanity . }\\
    \cdashline{2-2}[1pt/1pt]
    & \tabincell{l}{So man wants one day to establish a residence base on Mars and eventually emigrate to Mars and \\make it his second home .}\\
    \cdashline{2-2}[1pt/1pt]
    & \tabincell{l}{So man hopes one day to build a living base on Mars and eventually emigrate to make it a second \\home for humanity .} \\
    \cdashline{2-2}[1pt/1pt]
    & \tabincell{l}{So man wants to be able to build a living base on Mars and eventually emigrate to Mars , turning it \\into a second home . }\\
    \midrule
    \multicolumn{1}{l}{\multirow{10}{*}{\tabincell{c}{MixDiversity\\($\tau=0.35$)}}} 
    & \tabincell{l}{So . one day , humans want to build a living base on Mars and eventually emigrate to Mars and \\turn it into a second home .} \\
    \cdashline{2-2}[1pt/1pt]
    & \tabincell{l}{The human race , therefore , hopes that one day it will establish a residence base on Mars and \\ eventually immigrate to Mars to make it a second home .} \\
    \cdashline{2-2}[1pt/1pt]
    & \tabincell{l}{So man hopes one day to establish a base on Mars and eventually emigrate to Mars and turn it into \\a second home for man .} \\
    \cdashline{2-2}[1pt/1pt]
    & \tabincell{l}{Thus , mankind hopes that one day it will establish a living base on Mars and eventually immigrate \\to Mars , becoming a second home for humanity .}\\
    \cdashline{2-2}[1pt/1pt]
    & \tabincell{l}{So man wants to be able to build a residence base on Mars and eventually emigrate to Mars ,\\ making it thesecond home of man .} \\
    \bottomrule
    \end{tabular}
    
    \caption{Example outputs of BeamSearch and MixDiversity in WMT'17 \texttt{zh}$\rightarrow$\texttt{en}.}
    \label{tab:2}
\end{table*}
\end{CJK}

\section{Case Study}
In Table \ref{tab:2}, we illustrate a case of outputs from the MixDiversity and the BeamSearch in WMT'17 \texttt{zh}$\rightarrow$\texttt{en}. For the MixDiversity, we show the translation results under different $\boldsymbol{\tau}$. When $\boldsymbol{\tau}=0.15$, the 5 outputs of the MixDiversity follow a similar sentence pattern ``So man/human hopes one day to ...''. When the value of $\boldsymbol{\tau}$ increase from $0.15$ to $0.35$, both the number of sentence pattern and the number of subjects in the 5 generated translations are expanded and the differences between translations also becomes more obvious.

\section{Evaluation Results of the BERT-Score}
As aforementioned, reference BLEU and pairwise BLEU have been used to measure faithfulness and diversity in this work. However, BLEU simply counts
n-gram overlap between the inference and the reference, which can not account for meaning-preserving lexical and compositional diversity, e.g., synonyms and paraphrases. In contrast, the BERT-Score \citep{Zhang20iclr} seems to be a better measure, which computes a similarity score for each token in the inference sentence with each token in the reference sentence and correlates better with human judgments.

We apply the BERT-Score to evaluate the performance of different methods in WMT'14 \texttt{en}$\rightarrow$\texttt{de}, as shown in Tabel \ref{tab:bert-score}. we adopt the average BERT-score with reference (denoted as rf-BERTscore) to measure the faithfulness and the average pairwise BERT-score among generated sentences (denoted as pw-BERTscore) to measure the diversity. At last, we calculate the EDA using the BERT-Score (denoted as EDA-BERTscore) by substituting the BLEU score with the BERT-Score. We can see that the  MixDiversity (w/o Mixup training) gets the best pw-BERTscore and the best EDA-BERTscore.
\end{document}